\documentclass[journal,onecolumn]{IEEEtran}
\usepackage{cite}
\usepackage{amsmath,amssymb,amsfonts, pifont, bm}
\usepackage{algorithm, algpseudocode}
\usepackage{graphicx}
\usepackage{textcomp}
\usepackage{pgfplots}
\usepgfplotslibrary{external}
\usepackage{hhline}
\usepackage{xcolor}
\usepackage{url}
\renewcommand{\baselinestretch}{1.5} 

\begin{document}

\title{Bluetooth Low Energy Dataset Using In-Phase and Quadrature Samples for Indoor Localization}
\author{Samuel G. Leitch, \IEEEmembership{Student Member, IEEE}, Qasim Zeeshan Ahmed, \IEEEmembership{Member, IEEE}, Ben Van Herbruggen, Mathias Baert, Jaron Fontaine, Eli De Poorter, Adnan Shahid, \IEEEmembership{Senior Member, IEEE}, Pavlos I. Lazaridis, \IEEEmembership{Senior Member, IEEE}}




\maketitle
\begin{abstract}
One significant challenge in research is to collect a large amount of data and learn the underlying relationship between the input and the output variables. This paper outlines the process of collecting and validating a dataset designed to determine the angle of arrival (AoA) using Bluetooth low energy (BLE) technology. The data, collected in a laboratory setting, is intended to approximate real-world industrial scenarios. This paper discusses the data collection process, the structure of the dataset, and the methodology adopted for automating sample labeling for supervised learning. The collected samples and the process of generating ground truth (GT) labels were validated using the Texas Instruments (TI) phase difference of arrival (PDoA) implementation on the data, yielding a mean absolute error (MAE) at one of the heights without obstacles of $25.71^\circ$. The distance estimation on BLE was implemented using a Gaussian Process Regression algorithm, yielding an MAE of $0.174$m.
\end{abstract}


\section{Introduction}
Indoor Positioning (IP), has the potential to enable several different technologies that could massively improve patient management in care homes, asset management and general automation in warehouses, automated inspection and maintenance, etc~\cite{rrifloc, weco-slam}. IP assists in determining the device location by measuring location-dependent phenomena. These measurements could include the received signal strength (RSS) of wireless signals at the device's location, the angle of arrival (AoA) of wireless signals between the device and a base station, or the time of arrival (ToA) or time difference of arrival (TDoA) of wireless signals~\cite{Li2021}.

A wide range of wireless technologies are employable for RSS, AoA, ToA, and TDoA measurements. These include, but are not limited to, Bluetooth low energy (BLE)~\cite{Dinh-2020,Hind-2016}, ultra-wideband radio (UWB)~\cite{Fuhu-2023,Ahmed-2008,Ahmed-2014}, wireless fidelity (WiFi)~\cite{Wang2020_CSI,Ryan-2017} and millimeter (mm) wave radio~\cite{Amjad-2023,Osama-2017,Nair-2016}. UWB radio provides extremely accurate ToA and TDoA measurements due to the narrow width of its signals in the time domain, and more recently work has started on applying UWB to the task of AoA determination~\cite{iGentUWBAoA}. However, when considering a large-scale IP system, it is essential to consider the deployment cost of the technology. For this, BLE offers significant advantages due to the extremely low unit cost of individual devices and its presence in a large portion of modern consumer electronics such as smartphones. Its low power consumption when transmitting and receiving signals further enhances its appeal, making BLE an attractive candidate.

Determination of models requires a colossal amount of high-quality, labeled data for the underlying algorithms to accurately learn the relationship between the input and the desired output~\cite{Chen2020_IMU}. The collection and labeling of these datasets can be an extremely labor-intensive and time-consuming task, especially if these datasets are required to account for new environments, ranges of different variables, and possibly even different hardware types~\cite{Schwarzbach2022}. Consequently, it is vital to collect and share a wide variety of datasets from diverse environments within the research community. This will expedite the research and development process in this potentially revolutionary field.
\subsection{Related Works}
The authors of~\cite{DS_Girolami} provide a dataset for indoor localization utilizing the u-blox explorer (XPLR)-AoA-1, which does not expose the raw in-phase and quadrature (IQ) samples. Instead, it implements the algorithms for AoA determination internally in its firmware. In~\cite{DS_Girolami}, they use the azimuth and elevation angles calculated by the device to construct a dataset containing values gathered from four beacons distributed around a room. The ground truth (GT) included in the dataset consists of the $x$- and $y$-coordinates of the beacon. This dataset is intended to facilitate experiments on the triangulation technique. They include a variety of experiments, including continuous motion of the beacon and occasional periods of stopping. Finally, different localization experiments and the measured localization errors are reported for these scenarios. However, they do not mention the impact of the height that the beacon or anchors are held at, nor do they provide specifics about the distance between the beacon and the anchors.

The dataset collected in~\cite{DS_Maus} utilizes an $8$-element array arranged in a uniform linear array (ULA) configuration. This configuration is employed in two different scenarios: device-free respiration monitoring~\cite{Maus_respiration}, and indoor localization, as detailed in~\cite{Maus_localization}. Both datasets collect the raw IQ samples. The datasets include samples from scenarios where obstacles were present in the measuring space. However, the inclusion of any other variables, such as variable relative height or relative orientation was not mentioned.

The authors of~\cite{DS_Piazzese} provide a fully simulated BLE AoA dataset that uses mathematical models to generate the expected IQ samples for an $8$-element non-uniform rectangular array (NURA). Two different kinds of signals are provided: a pure-tone sinusoid and a BLE $5.1$ signal, both of which can have interfering signals imposed upon them. Despite the simulation having many tuneable parameters including the signal-to-interfering ratio (SIR) and signal-to-noise ratio (SNR), there are only two experimental variables: whether the signal is a pure tone or a BLE $5.1$ signal, and the angle of incidence of the desired signal~\cite{DS_Piazzese}.

The dataset at~\cite{DS_Paulino} utilizes a custom $8$-element uniform circular array (UCA) based on the Nordic semiconductor nRF52811 microcontroller. They have utilized $4$ beacons and $21$ collection positions and at each position have collected $600$ packets~\cite{paper_Paulino}. As outlined in the paper, this dataset was collected in an outdoor environment with no obstacles and no changes in the device orientation around any of the coordinate axes. While they have varied the position of the receiver array, they have not changed any of the other experimental variables for the data collection~\cite{paper_Paulino}. The authors tested the efficacy of their designed antenna array and the AoA algorithms without explicitly intending to release the dataset for public use.
\subsection{Contributions}
This paper details the collection of BLE IQ samples using the Texas Instruments (TI) BOOSTXL-AOA hardware. The dataset is collected across a range of different distances and elevations. To emulate a true industrial scenario, the dataset is collected with the transmitter when stationary as well as in motion. For the dataset, we have included the RSS values and the receiving channel of the BLE packet, as well as the GT values for the beacon, including its position and orientation. The dataset also includes repetitions of the experiments at predefined distances with obstacles placed between the transmitter and the receiver. Additionally, there are experiments where the device is in motion, alternating between moving towards and away from the receiver. The dataset is collected with a wide range of variables, that will enable the training of accurate and broadly applicable models. Finally, the contributions of this paper are as follows:
\begin{enumerate}
    \item A dataset of IQ samples and GT position data was collected from three different scenarios at varying heights and distances, both with and without obstacles. This dataset will be released for public research use.
    \item For higher accuracy, a method of automating GT data collection and labeling using motion capture (MoCap) technology is utilized.
    \item Verification of the dataset and GT data generation through the AoA determination algorithm used by the hardware manufacturer.
    \item An example application of the dataset using Gaussian Process Regression to perform distance estimation using RSS measurements, demonstrating the applicability of this dataset to a wide range of BLE localization research.
\end{enumerate}

Table~\ref{tab:rel_works} summarizes the related datasets found in the literature. It also compares them to the dataset collected for this paper. It can be observed that this collected dataset offers more variables in terms of data, such as variable height and distance of the tag, as well as including experiments with obstacles, a feature that only one other dataset provides.

\begin{table}[!t]
    \centering
    \caption{Experimental parameters and comparison of available BLE AoA datasets.}
    \label{tab:rel_works}
    \begin{tabular}{|c|c|c|c|c|}
\hline
         Paper&Antenna&Hardware&Experimental&Obstacles \\
		&Type&&Parameters&Included \\
 		\hline
        \cite{DS_Girolami}&Planar&XPLR-AOA-1&2& No\\ 
\hline
         \cite{DS_Maus}&ULA&RSS21&2& Yes\\ \hline
         \cite{DS_Piazzese}&NURA&Simulation&2&No\\ \hline
         \cite{DS_Paulino}&UCA&Custom&1&No\\ \hline
         \textbf{This work}&ULA&TI AoA&4&Yes\\ 
         &&BoostXL&&\\ \hline
    \end{tabular}
    
\end{table}
The paper consists of the following structure: Section~\ref{sec:AoA} provides an overview of AoA and emphasizes the importance of high-quality angle estimates. Section~\ref{sec:hw} will provide an overview of the hardware used for data collection, namely the TI BOOST-XL-AOA. Section~\ref{sec:data_col} provides detailed information on how the measurement campaign was conducted, including the rationale behind selecting different variables. Section~\ref{sec:structure} describes the structure of the dataset, including the process used to extract the GT angle from the recorded GT position data of the tag. In Section~\ref{sec:validation}, details are provided on how the dataset has been validated by implementing the hardware manufacturer's AoA determination algorithm. Finally, Section~\ref{sec:example} provides an example application of the dataset in the form of a GPR model for distance estimation. The paper is concluded in Section~\ref{sec:conc}. 

\section{Angle of Arrival (AoA)}\label{sec:AoA}
AoA is the process of calculating the incident angle of a received signal. AoA is an essential component of many IP techniques, i.e. in a single-anchor IP, the location of the tag being localized is determined by the measured angle and distance estimate to that base station~\cite{Ahmed-2020}. For a system such as this, accurate angle estimates, e.g. below $1^\circ$ error, are essential for the final positioning accuracy. This is especially true if operating a single-anchor solution at longer ranges, where the error of an angle estimate becomes much more significant. This can be demonstrated using trigonometry, where for a given expected angle error, $E[\theta_e]$, at a distance $d$ from the tag, the resultant expected localization error is estimated as:
\begin{equation}
    E[x_e] \approx d \tan\left(E[\theta_e]\right),
    \label{eqn:single_anchor}
\end{equation}
where $x_e$ represents the localization error estimate and $\tan(\cdot)$ is the tangent function. In this case, the distance is assumed to be known; however, in real applications, this will not be the case. The accuracy of the angle estimates is also extremely important for other angle-based IP techniques.

The most fundamental approach used to calculate the AoA of a wireless signal is the phase difference of arrival (PDoA) method. This method uses the measured differences in phase of the incoming signal at individual antenna elements, to calculate the extra distance that the signal had to travel to reach the further antenna element. This extra distance, along with knowledge about the spacing of the antenna elements, can be used to calculate the angle of the incoming signal. 
Several other methods for AoA determination can be found in the literature, including the multiple signal classification (MUSIC) algorithm, the estimation of signal parameters via rotational invariance techniques (ESPRIT), and machine learning (ML) models. Each of these relies on gathering data from more than one antenna arranged in an array.
\subsection{Angle of Arrival (AoA) on Bluetooth Low Energy (BLE)}
The capability for BLE to measure AoA was added to the specification in $2019$ with the release of revision $5.1$\cite{Leitch-2023}. This revision added the constant tone extension (CTE) to the BLE protocol. BLE employs Gaussian frequency shift keying (GFSK), which modulates the transmitted symbol by varying the frequency of the signal. As CTE appends a series of ones at the end of a BLE packet, it produces a sequence of identical symbols, resulting in a single frequency, or a constant tone, being transmitted. The CTE appended to the end of a packet can range in length from $16\mu s$ to $160\mu s$ and is made up of three distinct periods: the guard period which is $4\mu s$ long and has no sampling take place for its duration, the reference period which is $8\mu s$ long, and the sampling period which makes up the remainder. In the reference period, there are $8$ sampling slots allocated on a single antenna. The samples gathered in this period are to be used for determining how the phase of the signal is progressing over time~\cite{ble5.3_spec}. The sampling period is made up of a series of alternating switching and sampling slots, which can be either $1\mu s$ long or $2\mu s$ long. The length of these slots directly affects how many samples are collected from each CTE.

\subsubsection{TI PDoA Algorithm}

TI provide their own implementation of the PDoA algorithm, optimized for use on microcontrollers. One significant modification they make is in the way it performs carrier frequency offset compensation. It does this without the use of the samples obtained in the reference period of the CTE. This is accomplished by comparing the measured phase at the current antenna to its value in the previous round of switching\cite{Mohaghegh2021}. They also provide calibration figures for each antenna element on each channel to ensure the angle estimate is accurate\cite{rtls_toolbox}. The TI algorithm calculates an angle estimate on each array. By comparing RSS values, the algorithm can determine the side of the device the signal originated from. This is referred to as the RSS selection criteria in this paper. The TI algorithm also utilizes a simple moving average of the angle estimates to smooth out any outliers and improve accuracy in the angle estimation process.

\begin{figure}[!t]
    \centering
    \includegraphics[width=0.75\linewidth]{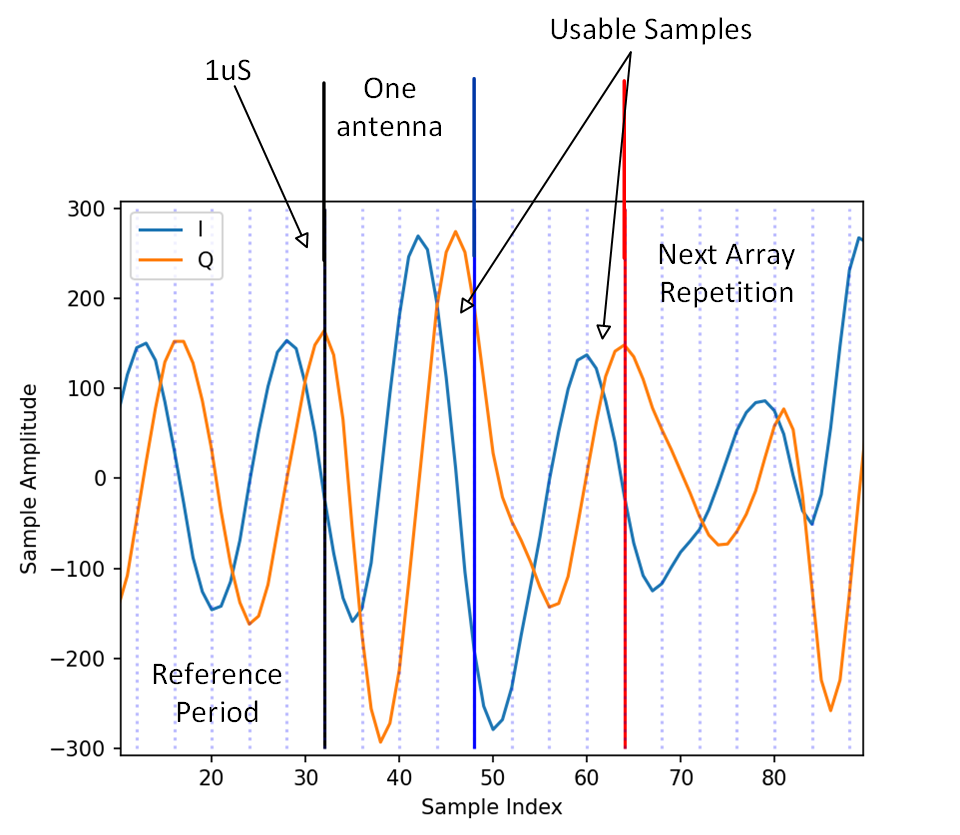}
    \caption{Example of an in-phase and quadrature (IQ) sample stream. }
    \label{fig:iq_processing}
\end{figure}
%
\section{Hardware}\label{sec:hw}
\begin{figure*}[!t]
    \centering
    \includegraphics[width=1.0\linewidth]{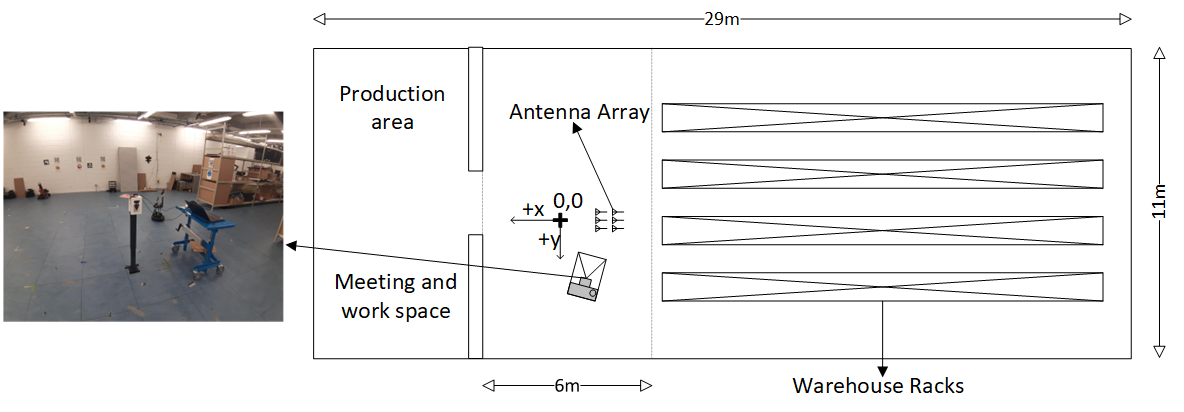}
    \caption{Layout of the experimental area. Experiments took place in a 6m x 11m area within a much larger room. Recreated from \cite{IIoTLab}}
    \label{fig:iiot-lab}
\end{figure*}
\begin{figure}[!t]
    \centering
    \includegraphics[width=0.75\linewidth]{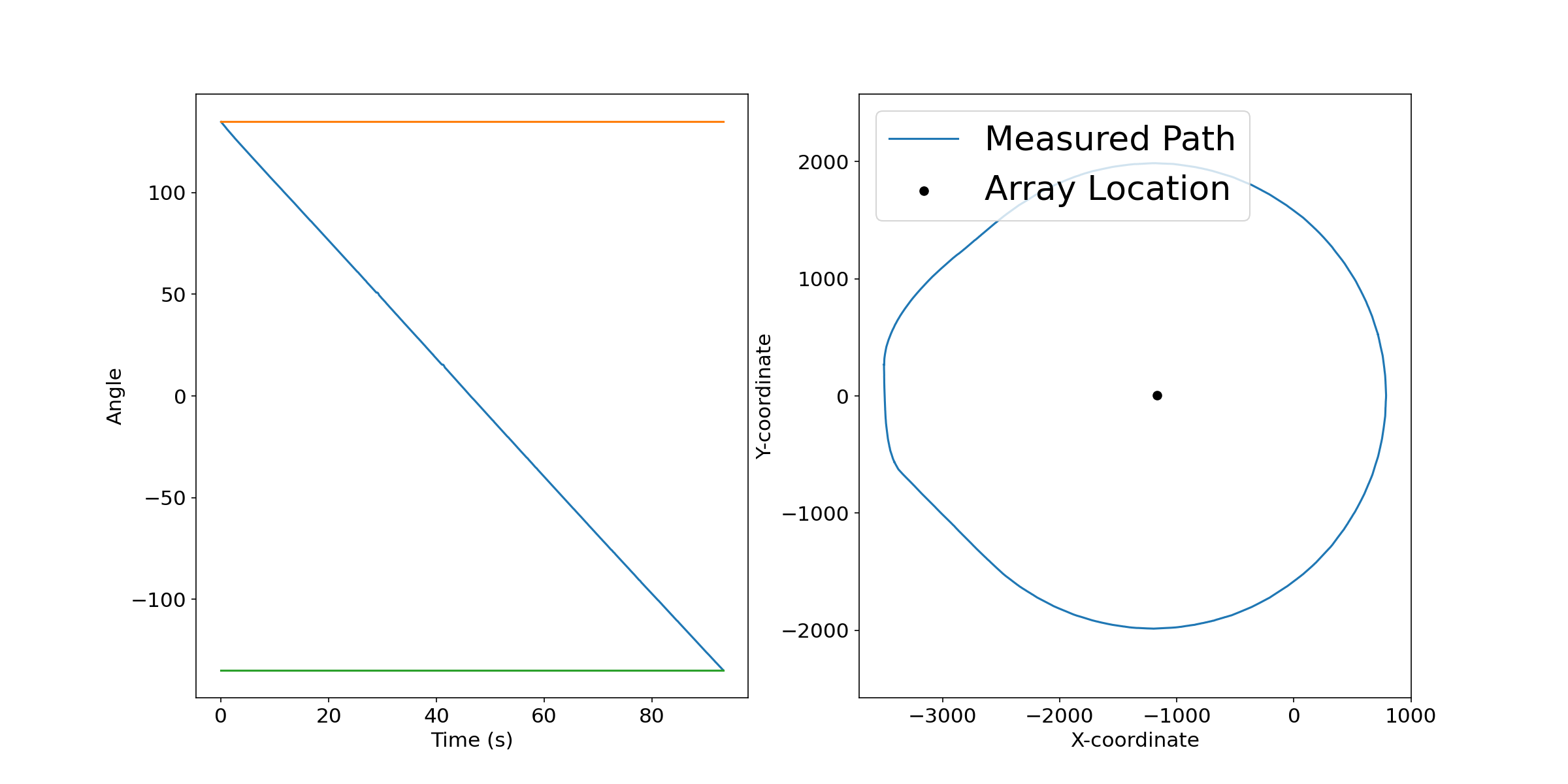}
    \caption{Analysis of the path traveled by the robot and the consequent GT angles for the three different experiment types. (a) Continuous}
    \label{fig:robot_paths-1}
\end{figure}

\begin{figure}[!t]
    \centering
    \includegraphics[width=0.75\linewidth]{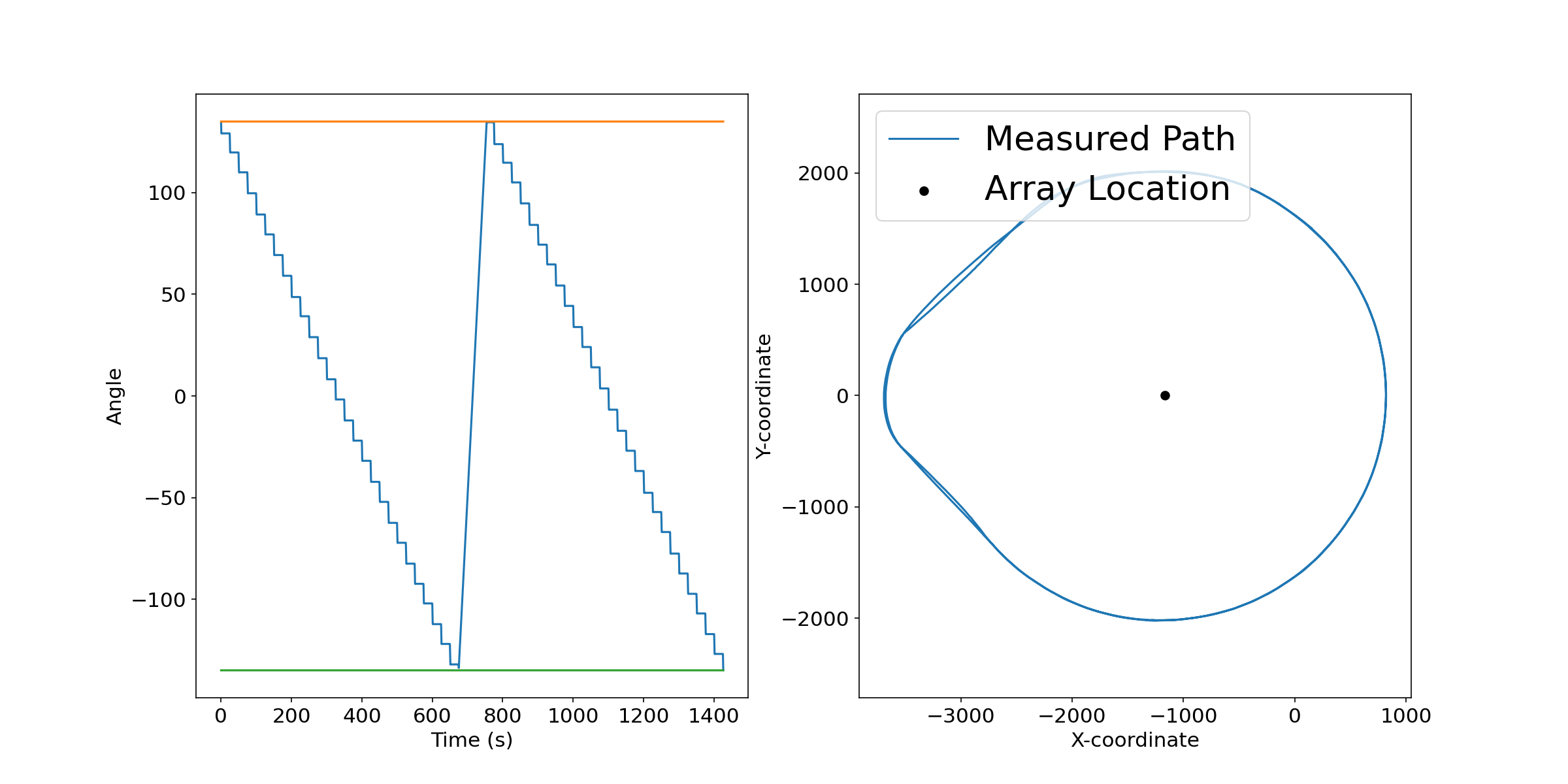}
    \caption{Analysis of the path traveled by the robot and the consequent GT angles for the three different experiment types.  (b) Stopping.}
    \label{fig:robot_paths-2}
\end{figure}

\begin{figure}[!t]
    \centering
    \includegraphics[width=0.75\linewidth]{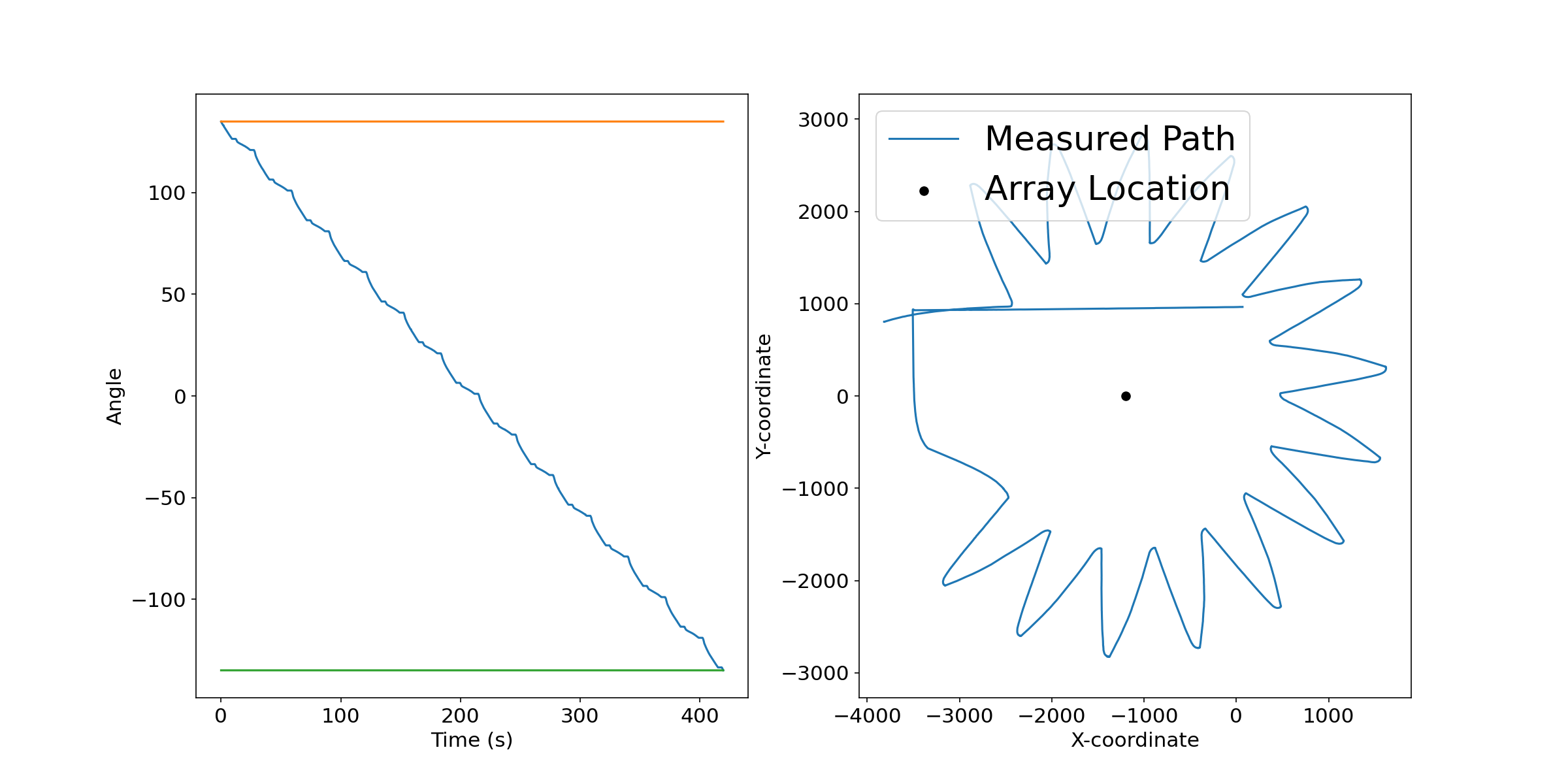}
    \caption{Analysis of the path traveled by the robot and the consequent GT angles for the three different experiment types. (c) Continuous zigzag.}
    \label{fig:robot_paths-3}
\end{figure}

The TI BOOSTXL-AOA provides two separate ULA which each contain three elements. These are arranged orthogonal to each other to provide a sensing range of $-135 ^\circ$ to $+135 ^\circ$. This sensing range arises from the fact that ULA is optimized for sensing in a $180^\circ$ range. One major downside of ULA is that they are incapable of distinguishing between angles that are separated by $180^\circ$\cite{diff_geom}. Three devices are required for a TI AoA system. This is because a hardware limitation in the TI devices means that the master cannot maintain an ongoing BLE connection while also performing IQ sampling. The three nodes in the TI AoA system are: A master node that initiates the BLE connection with CTE packets, a slave node that transmits the CTE packets, and a passive node that acts as a packet sniffer by passively receiving the packets of the BLE connection and performing the antenna switching before transmitting the data back to the computer. After flashing, only the passive and the master nodes need to be connected to the computer. The slave node can be battery-powered. Only the passive node requires an antenna array for operation, the rest can be a regular TI device. Due to the antenna switches, the AoA solution can only use $2\mu s$ time slots~\cite{rtls_toolbox}.

The TI BOOSTXL-AOA performs IQ sampling in an unusual way. As per the specification, during each sampling slot, a single sample is examined~\cite{ble5.3_spec}. However, the BOOSTXL-AOA performs sampling at a rate of $4$MHz throughout the CTE period. The filtering of the samples to ensure they are in the correct time slots is left to the user~\cite{rtls_toolbox}. A diagram of this filtering process can be observed in Fig.~\ref{fig:iq_processing}. The blue dotted lines represent $1\mu S$ slices, consisting of $4$ samples each. The black line marks the boundary with the reference period, the solid blue line represents the boundary between antenna $2$ and $3$ and the red line represents the boundary between array repetitions. With a $160\mu s$ CTE length, sampled at $4$MHz, the number of samples a user would expect from a single CTE, minus the $4\mu s$ guard period, would be $624$ IQ pairs. However, the maximum that ever gets received from the TI hardware is $511$, which is equivalent to $128\mu s$. This is because of the hardware constraints imposed by the size of the register that TI uses for storing IQ samples\cite{rtls_toolbox}.
\section{Data Collection Strategy}\label{sec:data_col}
This dataset is intended to be used for industrial robotics applications i.e. asset retrieval in warehouses or pipe inspection. These applications were the driving factor in the choices of environment and paths for the tag to follow in the data collection part of the process. 
\subsection{Measurement Environment}
The experiments were conducted in the IIoT lab at the iGent building of Ghent University. This is a $319m^2$ room that includes a large open section designed for drone localization and line of sight (LoS) experiments, as well as an area with metal racking similar to those found in warehouses\cite{IIoTLab}. A diagram of the layout of the IIoT lab can be found in Fig.~\ref{fig:iiot-lab}. The antenna array was mounted to a pole in the center of the $66m^2$ drone testing area of the lab. Although no other BLE devices were operating in the location at the time, UWB and WiFi devices were present. This, along with the metal racks, will help create an environment that closely approximates conditions found in an industrial setting.
\subsection{Motivation \& Measurement Protocol}
The BLE tag was attached to a small robot that could navigate itself around the room using a MoCap system implemented using $18$ cameras and Qualisys MoCap software capable of millimeter-level position accuracy. This robot was programmed to follow three different paths. The first path was a continuous sweep from $-135^ \circ$ to $+135^\circ $, maintaining the same distance (Fig.~\ref{fig:robot_paths-1}). This continuous path can be interpreted as the movement of a robot between fixed target points. The second was a sweep through the same angle range, maintaining the same distance, stopping for $25$ seconds every $5^\circ$  increment (Fig.~\ref{fig:robot_paths-2}). This scenario required two passes because the software controlling the robot was only technically capable of $10^\circ$ of separation between the points. By conducting two passes, it was possible to collect points with a granularity of $5^\circ$. This stopping path could be observed as the robots moving to a set location and taking measurements. The stopping paths serve a dual purpose, providing regularly spaced intervals where larger quantities of data are collected, providing rich data for the training of ML algorithms. For the third path, the robot would be in continuous motion again and sweep through the same angles; however, it would also vary the distance. This has been referred to in the dataset as a ``zigzag" path (Fig.~\ref{fig:robot_paths-3}). The zigzag paths provide a more challenging and more realistic version of the continuous case, with the tag changing in relative orientation throughout the path of the robot. 

The goal of the dataset is to provide a variety of data within the [$-135^\circ$,  $135^\circ$] range. To accomplish this, three different tag heights (800, 1100, and 1400mm above the ground) and four different tag distances (1500, 2000, 2500, and 3000mm between the transmitter and the receiver) have been chosen. The goal was to enable the training of ML models capable of generalizing to unobserved environments. Examples of the paths the autonomous robot followed for each of the experiment types and the resulting GT angles for a set distance and elevation can be observed in all figures (Figs. 3, 4 and 5) . The GT angles are defined for this figure as the azimuth angle. In this figure, the orange horizontal line represents the $+135^\circ$ mark, and the green horizontal line represents the $-135^\circ$ mark. The paths where the robot stops allowed for the collection of a multitude of data at specific angles for training, while the interstitial angles of the stopping experiments and the continuous movement experiments allowed for the collection of extra angles that can be used for training or testing and verifying the resulting algorithms for generalization. Also, in an attempt to make the dataset generalized to a real-world setting, the stopping and continuous experiment paths contain experiments where the path between the transmitter and receiver was impeded by obstacles as discussed in the upcoming section. The antenna array was configured with the default switching pattern, which switches between the three antennas in a single array for a given CTE. The slave node was configured with a connection interval of $100ms$, which affects the rate which CTE packets are sent to the master node.
\subsection{Measurement Obstacles}
\begin{figure}[!t]
    \centering
    \includegraphics[width=1.0\linewidth]{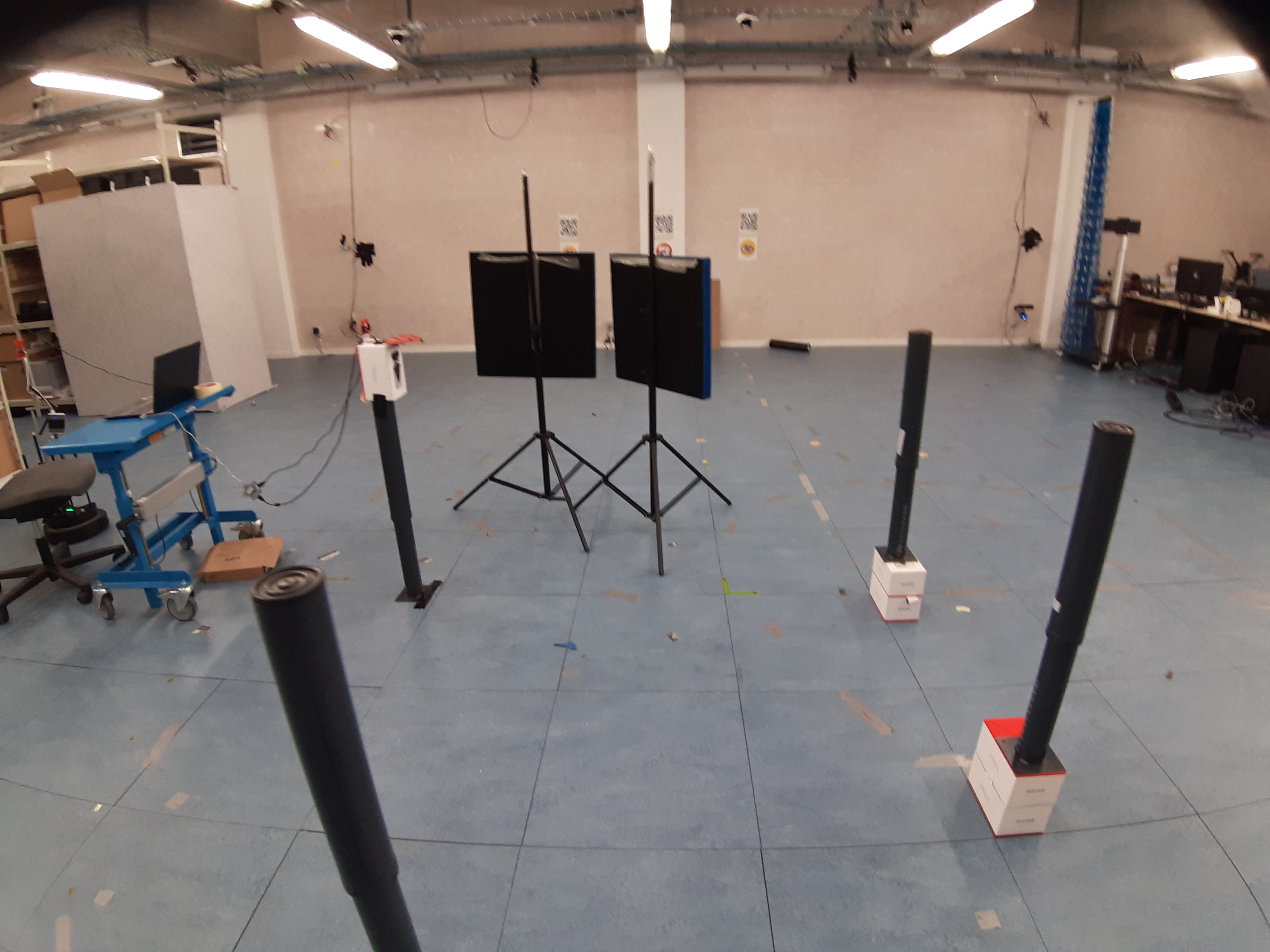}
    \caption{Types of obstacles placed between the transmitter and the receiver.}
    \label{fig:obstacles}
\end{figure}
The inclusion of experiments with obstacles between the tag and the receiver provides realistic examples of the stopping and continuous paths. In a subset of the experiments, obstacles in the form of reflective poles and absorbing material were placed in the LoS path between the transmitter and the receiver. These were to provide more varied and challenging data for training and testing robust ML models. Fig.~\ref{fig:obstacles} demonstrates examples of the kinds of obstacles placed in the LoS path. 
\section{Dataset Structure}\label{sec:structure}
The dataset is structured into a series of Javascript object notation (JSON) files. For each experiment, there is a pair of JSON files: one for the recorded signal data and one for the GT measurements. These are date and time stamped for when they were collected. A map of experiment names to JSON file names is located in the conversion dictionary. 
\subsection{Recorded Signal Data}
The signal data recorded in these files represents the features of the dataset. It was collected by serializing the TI real-time localization system. Packets were received from the TI data collection script into JSON and then written onto a file. Each entry is a dictionary containing metadata about the localization system used for data collection, a payload, and a local timestamp of the collected entry. It is in the payload that the recorded IQ samples can be found, along with the RSS of the packet and the channel that the packet was received. 

\begin{figure}
    \centering
    \includegraphics[width=1.0\linewidth]{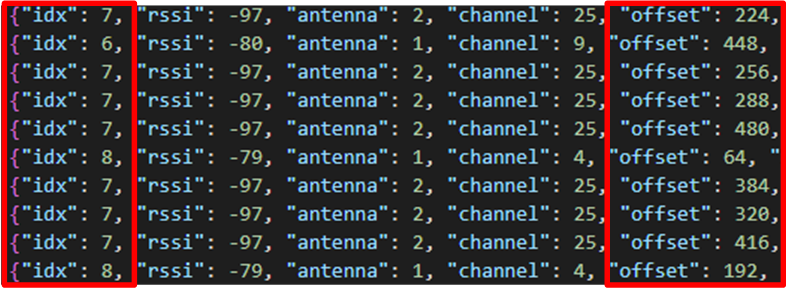}
    \caption{Example of a section of signal data file demonstrating UART packets being received out of order.}
    \label{fig:packet}
\end{figure}

The number of samples has been split into chunks of $32$ IQ pairs by the firmware on the antenna array before being transmitted to the data collection script through a universal asynchronous receiver/ transmitter (UART) connection. On the other end of the UART connection runs a Python script which processes the chunk and writes the results into a first in first out (FIFO) queue which in turn gets logged into the JSON file output. At the receiver, the order of the received chunks gets shuffled by the asynchronous nature of the data recording script.  As a result, the received chunks must be rearranged by the ``offset” key, highlighted on the right of Fig.~\ref{fig:packet}. An ``idx” key has also been added to the payload by editing the firmware of the device to enable grouping of the IQ chunks together according to whether they belong in the same CTE packet, highlighted on the left of Fig.~\ref{fig:packet}. When reordering the chunks, the “idx” key indicates whether IQ samples should be included in the same array of length $511$; and the “offset” key indicates where in the array of IQ samples to start writing the samples included in the chunk. The fully assembled CTE packet, along with its corresponding GT angle, constitutes one data point in the dataset.
\subsection{Ground Truth (GT) Measurements}
A specific GT for the angles has not been provided in the dataset, however, the measured position of the robot at a given timestamp is given instead. This has been measured using the MoCap system capable of a mm level of accuracy. Because the antenna array was positioned so that it was oriented to face along the $x$-axis of the room, we do not need to transform the robot coordinates into the antenna array reference frame. As a result, the azimuth angle can be calculated as:
\begin{equation}
    \Theta = \tan^{-1}\left(\frac{y_t - y_a}{x_t - x_a}\right),
    \label{eqn:GT_calculation}
\end{equation}
where $x_t$ and $x_a$ are the $x$-coordinates of the tag and anchor, and $y_t$ and $y_a$ are the $y$-coordinates of the tag and anchor, respectively. The position of the anchor has been measured in several experiments and found to be within $0.01$m of the position $x = -1200mm, y = 0mm$ and so the position of the anchor has been fixed to this position. If needed, calculation of the elevation angle can be achieved with the same formula as (\ref{eqn:GT_calculation}), replacing $y_t - y_a$ with $z_t - z_a$ and replacing $x_t - x_a$ with $\sqrt{x^2 + y^2}$. For the calculation of the elevation angle, the antenna array height has been fixed to $1100$mm.

When utilizing the position of the autonomous robot for calculating the GT azimuth angle, an issue arises when considering the update rates of the two systems. The antenna array at the passive node which performs the IQ sampling was configured with a connection interval of $100$ms, meaning a new stream of IQ samples is received at that same interval, while the MoCap system for the position update of the robot updates at a mean rate of $57.8$ms, measured across all of the experiments. This means that very few of the CTE reception times directly correspond to a GT reception time. While it would be possible to simply use the nearest GT position update, this would not provide the level of GT angle accuracy necessary for a valid dataset. Therefore, it was decided that in the instances where an exact timestamp match between CTE and GT was not available, the two position updates on either side of the CTE timestamp would be found and linear interpolation would be performed on the GT positions. 
\begin{table*}
    \centering
    \caption{Report of angle estimates for scenarios a) and b), comparing the impact of obstacles on the angle estimates.}
    \label{tab:obstacle_impact}
    \begin{tabular}{|c|c|c|c|c|c|c|}\hline
         &\multicolumn{2}{|c|}{MAE ($^\circ$)}&\multicolumn{2}{|c|}{Range MAE ($^\circ$)}&\multicolumn{2}{|c|}{Moving Average MAE ($^\circ$)}  \\
         Height (mm)&Without obstacles&With obstacles&Without obstacles&With obstacles&Without obstacles&With obstacles\\\hline
         800&33.04&36.34&12.30&16.29&28.26&30.74\\\hline
         1100&25.71&26.29&10.45&11.29&21.87&22.18\\\hline
         1400&39.26&44.58&17.13&25.89&33.04&37.41\\\hline
    \end{tabular}
\end{table*}
It can be seen from Fig.~\ref{fig:robot_paths} that the range of angles between $-135^\circ$ and $135^\circ$ has been exhaustively cataloged, with the zigzag path providing extra orientations of the robot concerning the antenna array. These orientations have also been cataloged and provided in the GT data files in the form of a three-dimensional ($3$D) rotation matrix where the columns represent the local coordinate axes in the global reference frame\cite{qualisys_rotations}. These extra orientations could be used to research pose-invariance for AoA determination, which can prove to have an effect on the accuracy of an AoA-system~\cite{aoa_orientation}. A full detailing of the statistics of the data, including the number of data points collected in each experiment, can be found in the dataset repository The majority of the data points were collected in the stopping experiments to provide the bulk of the samples for the training of ML models. The angles between the stopping points, as well as the continuous, zigzag, and obstacle experiment sets, were included for testing of the resulting models.
\section{Data Validation}\label{sec:validation}
The gathered dataset has been evaluated using the TI PDoA implementation, which was ported from the C-code firmware over to Python for post-processing. This approach provided us with two benefits i) it enabled us to assess the quality of the collected dataset, and ii) it established a baseline for future work by employing the performance of the TI PDoA algorithm as a benchmark for comparison for future work on improving AoA determination algorithms.
\begin{figure}
    \centering
    \includegraphics[width=1.0\linewidth]{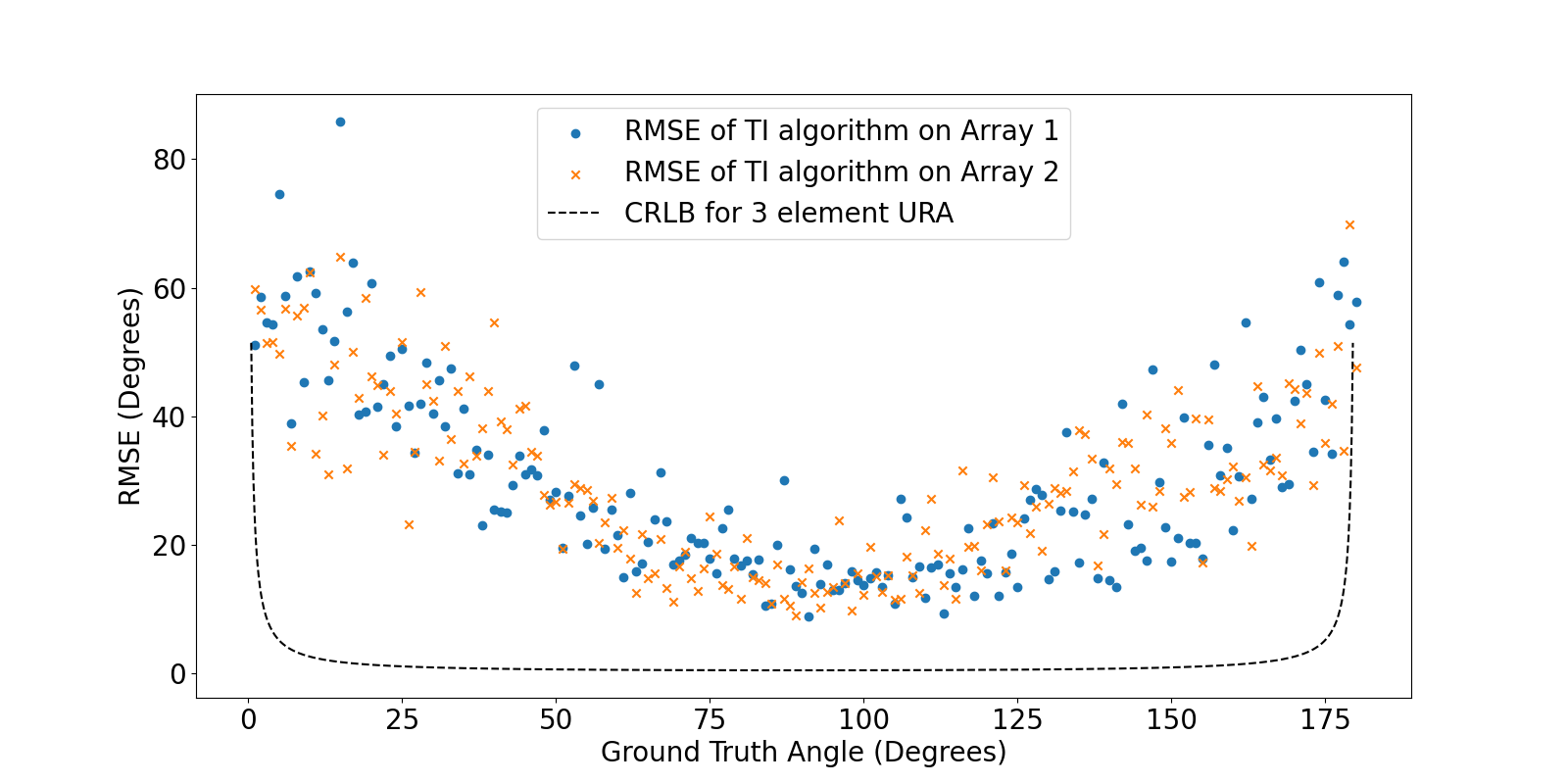}
    \caption{Comparison of angle estimates from the TI algorithm with the Cramer-Rao Lower Bound}
    \label{fig:crlb}
\end{figure}
\begin{figure}
    \centering
   \includegraphics[width=0.75\linewidth]{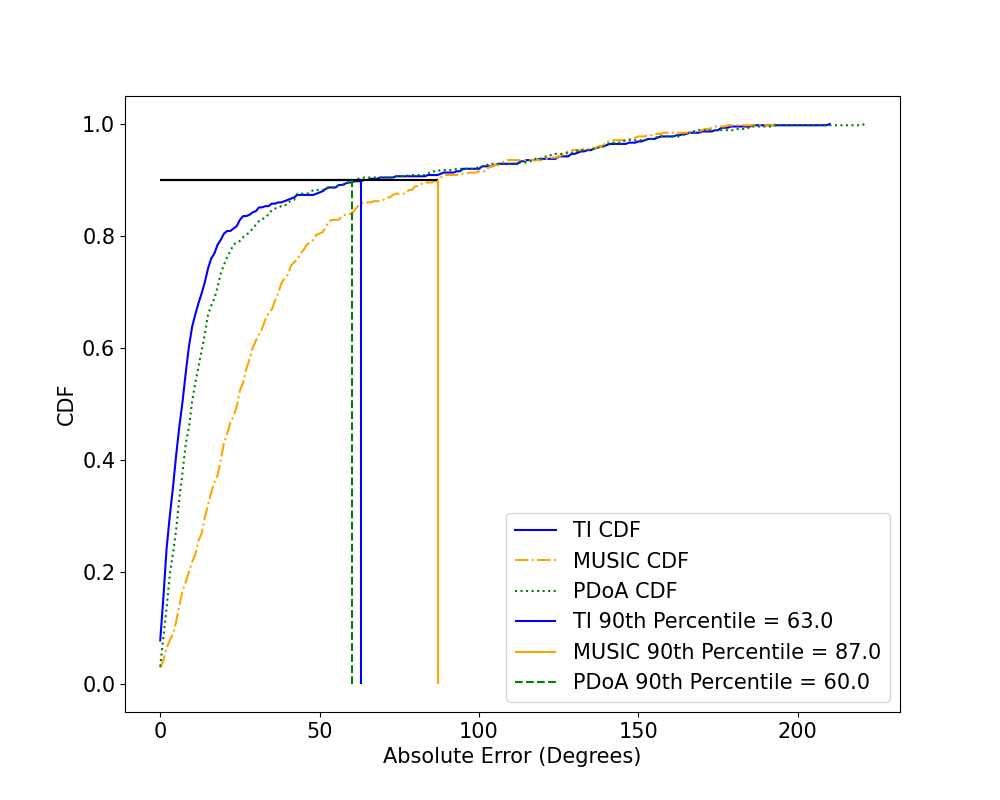} 
\caption{CDF generated by applying the TI, MUSIC, and PDoA algorithms to the IQ samples on three experiments at the same tag height. (a) is continuous} 
\label{fig:1100_cont}
\end{figure}
\begin{figure}
    \centering
\includegraphics[width=0.75\linewidth]{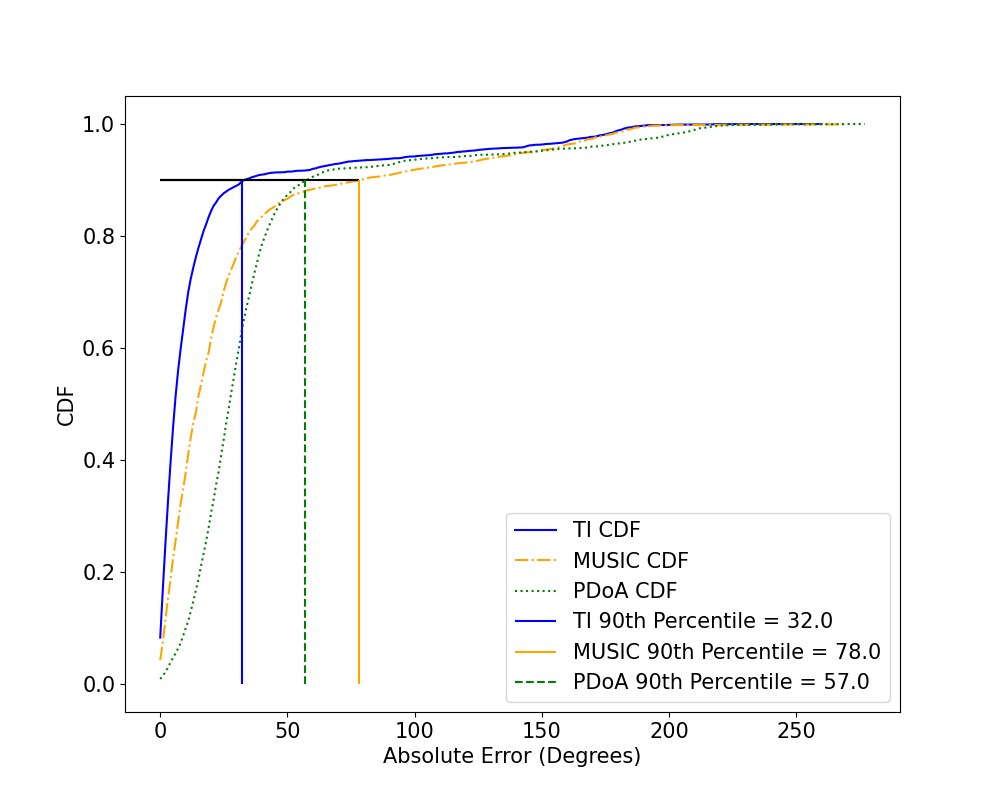}    
\caption{CDF generated by applying the TI, MUSIC, and PDoA algorithms to the IQ samples on three experiments at the same tag height. (b) is stopping}\label{fig:1100_stop}
\end{figure}
\begin{figure}
    \centering
\includegraphics[width=0.75\linewidth]{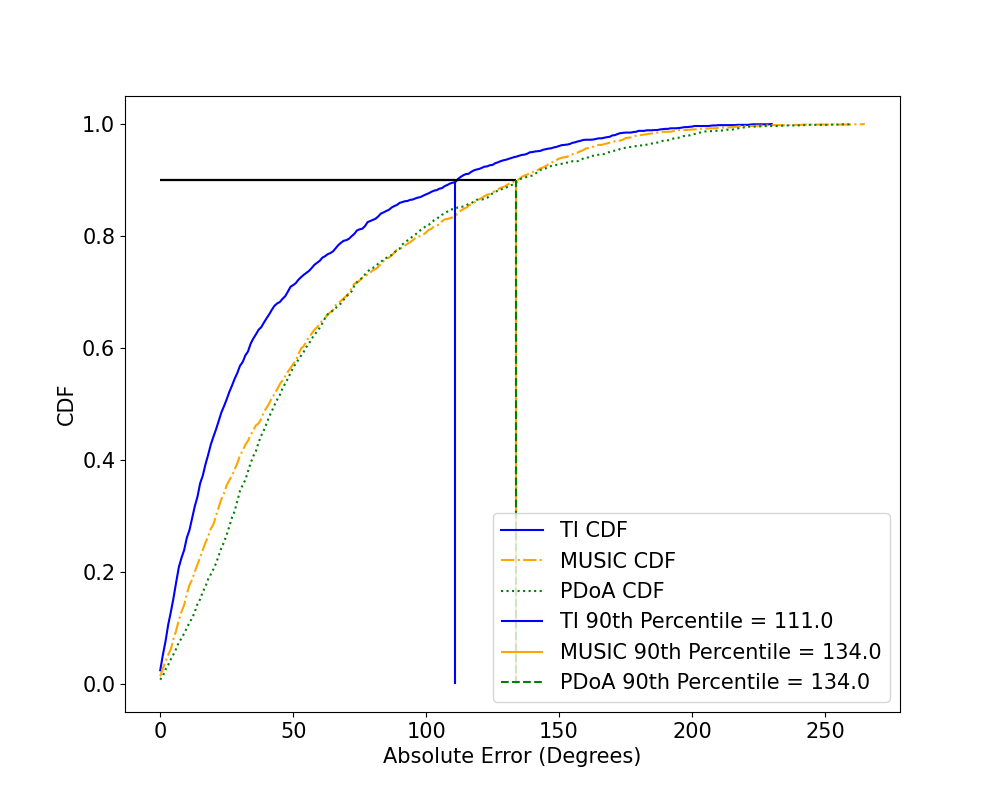}  
\caption{CDF generated by applying the TI, MUSIC, and PDoA algorithms to the IQ samples on three experiments at the same tag height. (a) is continuous, (b) is stopping, and (c) is continuous zigzag.}
    \label{fig:verification_1100}
\end{figure}
Fig.~\ref{fig:crlb} shows a plot of the angle estimates from the individual antenna arrays against the GT angle. This plot was generated by dividing the dataset based on which side of the array the packet was transmitted from, filtering out any instances where the packet had to travel across the width of the device before being received. The GT angles were then transformed from the full angle range of $[-135^\circ, 135^\circ]$ into the angle range for a ULA $[0^\circ, 180^\circ]$. For example, for angle estimation on the first array in the device:
\begin{equation}
    \theta_{ULA} = \theta_{Array1} + 45,
\end{equation}
where $\theta_{ULA} \in [0^\circ, 180^\circ]$ and $\theta_{Array1} \in [-45^\circ, 135^\circ]$. Following this, the TI AoA algorithm was applied, without the RSS selection criteria. This gave the fairest possible comparison between the performance of the algorithm against the lower-bound, which was calculated using the formula\cite{crlb_ble}:
\begin{equation}~\label{eq-4}
    CRLB\left(\theta\right) = \sqrt{\frac{3}{2pmn\left(n-1\right)\left(2n-1\right)}}\frac{\sigma c}{\pi df_c\sin\theta},
\end{equation}
where $\theta$ is the GT angle, $f_c$ is the carrier frequency, $d$ is the spacing between receiving elements, $c$ is the speed of light, $\sigma$ is the standard deviation of the received phases, $n$ is the number of receiving elements, $p$ is the number of samples taken in a single sampling slot and $m$ is the number of repetitions through the switching cycle that occur for a single CTE. (\ref{eq-4}) calculates the lower bound on the RMSE of the angle estimates and the RMSE is calculated as follows:
\begin{equation}
    RMSE\left(\hat{\theta}\right) = \sqrt{\overline{\left(\theta - \hat{\theta}\right)^2}},
\end{equation}
where $\hat{\theta}$ is the estimated angle. 
\subsection{Impact of Obstacles}
Table \ref{tab:obstacle_impact} contains the performance results of the TI AoA determination algorithm on the gathered dataset for scenarios a) and b) averaged over the different distances at each of the heights. The results have been broken down based on whether or not the experiment contained obstacles between the tag and the array. They have been further broken down by the height of the tag, to provide a level of granularity for analysis. Three metrics have been used for the table: the overall MAE of the experiments calculated as:
\begin{equation}
    MAE(\bm{\Theta}) = \overline{\lvert\left(\bm{\Theta} - \hat{\bm{\Theta}}\right)\rvert},
\end{equation} 
where $\hat{\bm{\Theta}}$ is the vector of all predicted angles and $\bm{\Theta}$ is the vector of all corresponding GT angles. Range MAE is the MAE for the region $\Theta \in [-50^\circ, 50^\circ]$. This metric was included for reasons described in detail in Section VI.C. A window size of $6$ is used to calculate the moving average MAE. It was included as it was used by TI in their firmware implementation. Looking at Table \ref{tab:obstacle_impact}, the algorithm's performance in the range MAE category is good, with the obstacles making a measurable difference to the performance. This can be interpreted as a justification for including experiments with obstacles in the dataset, as it provided a more challenging environment for training robust ML models.
\subsection{Impact of Scenario}
Figure \ref{fig:verification_1100} shows plots of the performance of three different AoA determination algorithms. The algorithms are: the TI algorithm which was ported from the device firmware into python for data exploration; the MUSIC algorithm which calculates the covariance matrix of the input data before performing eigenvalue decomposition; and the PDoA algorithm. It can be observed that in all three scenarios the TI algorithm performs on par, if not better, than the other two algorithms. Performances in the continuous and stopping experiment classes are very reasonable, with around $80\%$ of the errors being less than $50^\circ$. The degradation of performance in the zigzag category is expected and can be explained by the variable orientation of the tag which was designed into the experiment set. The shapes of the CDF in Fig.~\ref{fig:verification_1100} imply the existence of extreme outliers in the angle estimates. While some of these will have been caused by the performance of the algorithm and the array, as evidenced by Fig.~\ref{fig:crlb}, the most extreme of these outliers will have been caused by fluctuations in the RSS. The impact of the RSS on the final angle estimate is explained in more detail in the next sub-section.
\subsection{Impact of RSS}\label{sec:rss_impact}
\begin{figure}[!t]
    \centering
    \includegraphics[width=1.0\linewidth]{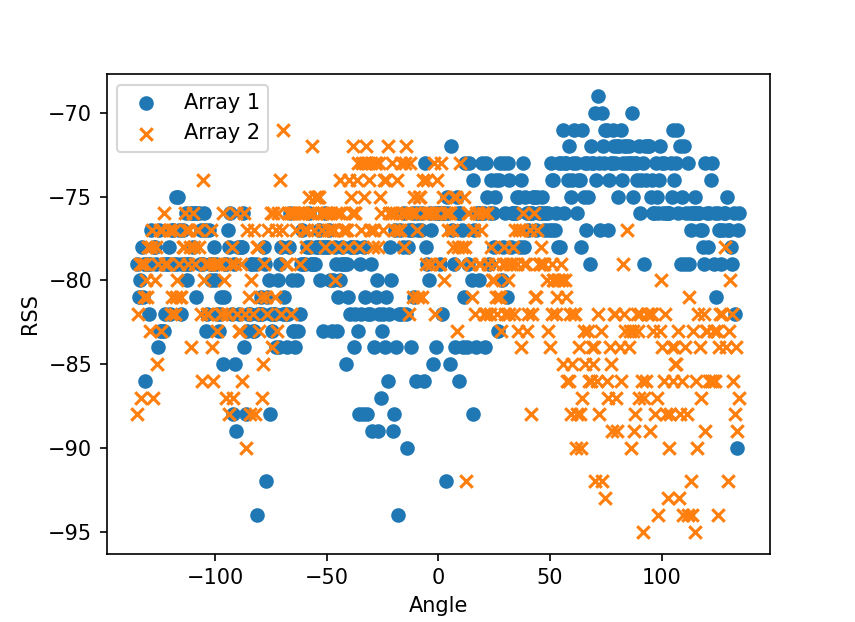}
    \caption{Plot of the RSS for each sub-array for each angle measured in a continuous experiment at $2$m distance and a height of $1.1$m.}
    \label{fig:experiment_rss}
\end{figure}

The TI PDoA algorithm calculates phase differences between three different pairs of antennas in the array: antenna $1$ and $2$, antenna $2$ and $3$, and antenna $1$ and $3$, respectively. The algorithm calculates the average phase difference for each of these pairs, determines the corresponding AoAs, and then makes final adjustments to the output angle based on the specific array configuration and calibration measurements conducted by TI in a laboratory setting~\cite{rtls_toolbox}. This process is repeated for the other array on a separate CTE packet and the final angle estimate is selected based on which of the packets has a higher RSS value. Fig.~\ref{fig:experiment_rss} is a plot of the RSS of the CTE packets plotted against the measured GT angle of the tag. As it shows, this selection criteria serves as an effective measure for deciding which side of the device the tag is on, with higher RSS values at positive angles corresponding to higher RSS on Array 1. However, it is far from perfect. The left-hand side of Fig.~\ref{fig:experiment_rss} plots the negative GT angles which show a clear deviation from this expected behaviour. This deviation in behaviour is consistent across the experiments. Future tests in the same environment will help establish whether this behaviour is caused by environmental factors or if it was a hardware issue at the time of the experiments. Regardless, this serves to strengthen the dataset as it provides realistic and unpredictable data. Fig.~\ref{fig:crlb} also demonstrates that individual angle estimates themselves are unaffected, with the only effect being to select the incorrect estimate after estimates have been made on both arrays. To refine the final angle estimate, TI maintains a moving average of the previous $6$ angle estimates and employs the averaged value as the final angle estimate. This same process of RSS selection and moving average was also applied to the PDoA and MUSIC algorithms as observed in Figs.~\ref{fig:verification_1100} and \ref{fig:verification_1400}.

\begin{figure}[!t]
    \centering
    \includegraphics[width=1.0\linewidth]{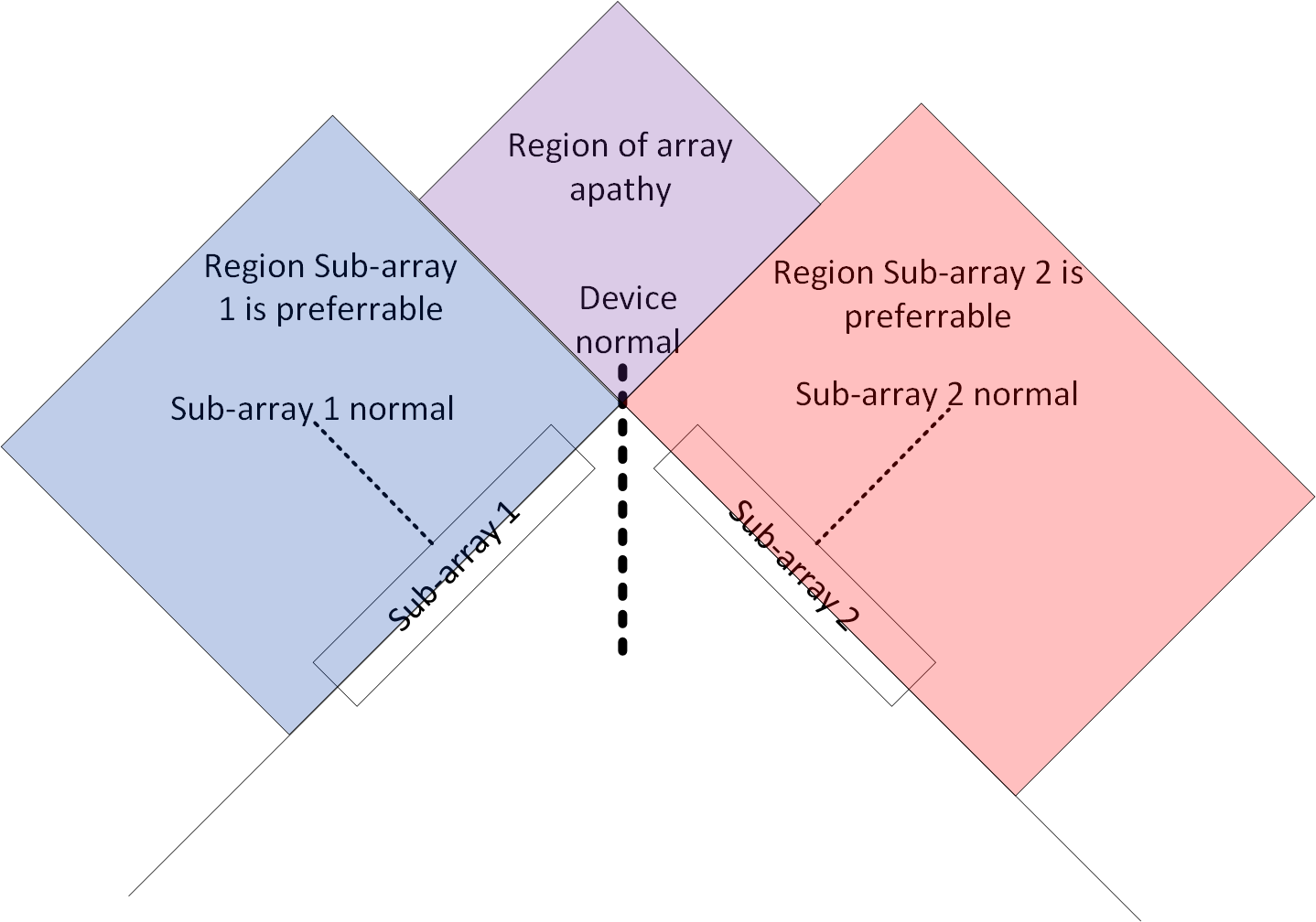}
    \caption{Diagram demonstrating the arrangement of sub-arrays and the corresponding region of overlap around $0^\circ$.}
    \label{fig:array_agnosticism}
\end{figure}

It is worth noting that, in the region surrounding $0^\circ$, angle estimates tend to be more accurate than those for the rest of the experiment. This observation is supported by the Range MAE column of Table \ref{tab:obstacle_impact}, where the MAE in the region of $[-50^\circ, 50^\circ]$ is consistently lower. Such a phenomena can be partially explained by the arrangement of the arrays. In Fig.~\ref{fig:array_agnosticism}, the two ideal detection regions for sub-arrays $1$ and $2$ are marked in red and blue, respectively. Each of the sub-arrays are capable of sensing in a region from $0^\circ$ to $180^\circ$, with their normal at $90^\circ$. Given that they are offset from each other by $90^\circ$, this means that there is a region of $90^\circ$ within which the estimate from either sub-array is likely to be accurate. This is true so long as the GT angle does not approach $0^\circ$ or $180^\circ$ because, as Fig.~\ref{fig:crlb} shows, the angle estimates at these limits are extremely inaccurate, which shows up on plots of the angle estimates as a degradation in estimate quality when the GT angle is at $0^\circ$. Consequently, degradation of RSS has a much smaller impact on the final estimate when in the region of $[-50^\circ, 50^\circ]$. This region has been dubbed the ``region of array apathy".

\subsection{Impact of Height}

Fig.~\ref{fig:verification_1400} is a repetition of Fig.~\ref{fig:verification_1100} with the tag at an elevation of $1.4$m. Comparing the two, the difference in height seems to have had a noticeable impact on the statistical distribution of the angle estimates. This is evidenced by the longer tails on the plots of the CDF. Comparing the rows of Table~\ref{tab:obstacle_impact} against each other seems to confirm this, with $1.1$m performing the best out of the three selected tag heights.

\section{Experiment}\label{sec:example}
This dataset is usable for more than just AoA determination experiments. To prove the versatility of this dataset, experiments on BLE range-finding will now be detailed. BLE range-finding is essential to two main localization techniques, namely the single-anchor and multilateration techniques. When discussing the importance of the accuracy of angle estimates for the single-anchor technique in Section \ref{sec:AoA}, the distance in (\ref{eqn:single_anchor}) was assumed to be known. Naturally, this isn't usually the case for a localization system, so the final error of an IP system based on the single-anchor technique relies heavily on the accuracy of the distance estimate. Unfortunately, range-finding on BLE using RSS has historically proven to be very inaccurate, even with multiple attempts to apply methods to improve it. This is due to many reasons, including the noisy and unpredictable nature of the RSS measurement as well as the susceptibility of the measurements to changes in the environment. Range-finding in BLE has traditionally been implemented using RSS measurements and the log-loss model to calculate the range through the following equation\cite{Zafari2019}:
\begin{equation}
    d(x) = 10^{-\left(\frac{x - x_0}{10\alpha}\right)},
    \label{eqn:range_finding}
\end{equation}
where $x$ is the measured RSS value, $x_0$ is the measured RSS at a set reference distance, and $\alpha$ is the path loss exponent which modifies the slope to account for environmental factors. The $\alpha$ value in (\ref{eqn:range_finding}) can vary not just with the environment and the arrangement of objects in it, but also with the channel that a packet is received on\cite{Leitch_nb}. This has been exploited previously, but the results are limited based on the performance of the log-loss model\cite{Nikodem2021}.

Gaussian Processes generalize the Gaussian distribution to be over an infinite set of candidate functions~\cite{Ahmed-2013,gpr_jadaliha,gpr_esteve} with mean $\mu\left(\mathbf{x}\right)$ and covariance $\mathbf{K}$ where $K_{ij} = k(\mathbf{X}_i, \mathbf{X}_j$), with $k$ being the covariance function, or kernel\cite{gpr_book, gpr_esteve}. The matrix $\mathbf{X} \in \mathbb{R}^{N\times 37}$ is the training set of $N = 1000$ RSS vectors randomly sampled from the dataset with one measurement per channel, where $37$ is the number of BLE data channels. In this example application the chosen kernel was a combination of the Radial Basis Function (RBF) and linear kernels with equations
\begin{equation}
    k_{RBF}(\mathbf{x}, \mathbf{x}') = \sigma_{RBF}^{2}\exp{\left(\frac{||\mathbf{x}-\mathbf{x}'||}{2\ell^2}\right)},
\end{equation}
\begin{equation}
    k_{linear}(\mathbf{x}, \mathbf{x}') = \sigma_{linear}\left(\mathbf{x}\mathbf{x}^T\right),
\end{equation}
where $\sigma_{RBF}^2$ is the variance for the RBF kernel, $\sigma_{linear}^2$ is the variance of the linear kernel, and $\ell$ is the lengthscale of the RBF kernel. Each of these can be tuned as hyperparameters of the model using Bayesian optimization. 

\begin{algorithm}
\caption{Process for generating the GPR dataset}
\label{alg:gpr_dataset}
    \begin{algorithmic}
        \Require $N$\Comment{Total number of entries in dataset}
        \Require $M$\Comment{Number of entries in GPR dataset}
        \Require $\mathbf{rss} \in \mathbb{R}^N$\Comment{Vector of all RSS values}
        \Require $\mathbf{d} \in \mathbb{R}^N$\Comment{Vector of all GT distances}
        \Require $\mathbf{t} \in \mathbb{R}^N$\Comment{Vector of all Timestamps}
        \Require $\mu_t$ \Comment{Mean timestep between dataset entries}
        \Require $\mathbf{c} \in \mathbb{R}^N$\Comment{Vector of all Channels}

        \State $i \gets 0$
        \State $j \gets 0$
        \State $\mathbf{F} \gets \mathbf{0}^{M\times 37}$\Comment{Feature matrix, initialized to 0}
        \State $\mathbf{l} \gets \mathbf{0}^M$\Comment{Labels vector, initialized to 0}

        \While{$i < N$}
            \If{$i > 0$}
                \State $dt \gets \mathbf{t}_i - \mathbf{t}_{i-1}$
                \If{$dt > \mu_t$}
                    $\mathbf{F}_j \gets \mathbf{0}^{1\times 37}$
                \EndIf
            \EndIf
            \State $rss \gets \mathbf{rss}_i$
            \State $c \gets \mathbf{c}_i$
            \State $\mathbf{F}_{jc} \gets rss$

            \If{$nonzero(F_j) == \emptyset$}\Comment{If all columns are filled}
                \State $\mathbf{l}_j \gets \mathbf{d}_j$
                \State $j \gets j + 1$
            \EndIf
            \State $i \gets i + 1$
        \EndWhile{}
            \end{algorithmic}
\end{algorithm}

Algorithm \ref{alg:gpr_dataset} outlines the process followed to extract the GPR dataset from the total dataset. This process assembled the training set in a manner that would closely resemble how it would be actually implemented on a real system, where packets are received sequentially with the channel number being drawn from a uniform distribution. The input vector requires one RSS value for each channel. Measurements made during the dataset creation process determined that the average number of packets required for each input vector was 50. This is because the random nature of the channel numbers caused duplicate entries, which had to be rejected. The dataset was assembled from data gathered in all of the experiments, including the experiments from the zigzag scenario. By putting a threshold on the timestep between each packet, the assembly of feature vectors from packets in different experiments was prevented. Following the dataset assembly, $1000$ rows of $\mathbf{F}$ and $\mathbf{l}$ were randomly drawn to form the training set. The value of $1000$ represents $~20\%$ of the total dataset and was chosen to balance approximating the posterior probability distribution against inference time\cite{Hensen-2013}. Testing of the model on the remaining samples in the dataset yielded an MAE of the distance estimates of $0.174$m, a median error of $0.08m$, and a variance of $0.07$m. In comparison, the Log-Loss (LL) model achieved an MAE of $12.35m$ and a median error of $1.786m$. By accounting for the variable channel by creating a separate LL model for each channel\cite{Nikodem2021}, Channel Informed Log-Loss (ChILL) is possible. This achieved an MAE of $4.13m$ and a median error of $1.44m$. Plots of the absolute estimation error for the GPR algorithm, alongside the CDF of the absolute error for GPR, LL, and ChILL can be found in Fig.~\ref{fig:gprcdf}. It should be taken into account the fact that, for a connection interval of $100ms$, requiring an average of 50 packets per input vector means that each position update would occur after $5s$. Means to reduce this must be considered.

\section{Conclusion}~\label{sec:conc}
This paper has detailed the collection and verification of a dataset of BLE IQ samples for use in indoor localization. Labelling of the samples was automated through the use of MoCap technology. The dataset and automated labeling were verified by applying the PDoA algorithm developed by TI to the data and demonstrating the performance. Comparisons have also been made between the performance of the TI PDoA algorithm and an implementation of the MUSIC algorithm and a more n\"ive implementation of PDoA. Of particular note is the performance in the region between $-50^\circ$ and $50^\circ$, where high variance in the RSS of the signal did not corrupt the final angle estimates. The mean MAE in this region being $15.42^\circ$ for the TI PDoA algorithm. These results provide a baseline for comparison for future work on indoor localization using BLE. As an example application of the dataset, a GPR algorithm was developed for distance estimation on BLE RSS measurements, which yielded an MAE of $0.174m$.

\bibliography{IEEEfull,refs}

\begin{thebibliography}{10}
\providecommand{\url}[1]{#1}
\csname url@samestyle\endcsname
\providecommand{\newblock}{\relax}
\providecommand{\bibinfo}[2]{#2}
\providecommand{\BIBentrySTDinterwordspacing}{\spaceskip=0pt\relax}
\providecommand{\BIBentryALTinterwordstretchfactor}{4}
\providecommand{\BIBentryALTinterwordspacing}{\spaceskip=\fontdimen2\font plus
\BIBentryALTinterwordstretchfactor\fontdimen3\font minus
  \fontdimen4\font\relax}
\providecommand{\BIBforeignlanguage}[2]{{%
\expandafter\ifx\csname l@#1\endcsname\relax
\typeout{** WARNING: IEEEtran.bst: No hyphenation pattern has been}%
\typeout{** loaded for the language `#1'. Using the pattern for}%
\typeout{** the default language instead.}%
\else
\language=\csname l@#1\endcsname
\fi
#2}}
\providecommand{\BIBdecl}{\relax}
\BIBdecl

\bibitem{rrifloc}
S.~Deng, W.~Zhang, L.~Xu, and J.~Yang, ``Rrifloc: Radio robust image
  fingerprint indoor localization algorithm based on deep residual networks,''
  \emph{IEEE Sensors Journal}, vol.~23, no.~3, pp. 3233--3242, 2023.

\bibitem{weco-slam}
V.~Kachurka, B.~Rault, F.~I. Ireta~Muñoz, D.~Roussel, F.~Bonardi, J.-Y.
  Didier, H.~Hadj-Abdelkader, S.~Bouchafa, P.~Alliez, and M.~Robin,
  ``Weco-slam: Wearable cooperative slam system for real-time indoor
  localization under challenging conditions,'' \emph{IEEE Sensors Journal},
  vol.~22, no.~6, pp. 5122--5132, 2022.

\bibitem{Li2021}
Y.~Li and K.~Yan, ``Indoor localization based on radio and sensor
  measurements,'' \emph{IEEE Sensors Journal}, vol.~21, no.~22, pp.
  25\,090--25\,097, 2021.

\bibitem{Dinh-2020}
T.-M.~T. Dinh, N.-S. Duong, and K.~Sandrasegaran, ``Smartphone-based indoor
  positioning using ble ibeacon and reliable lightweight fingerprint map,''
  \emph{IEEE Sensors Journal}, vol.~20, no.~17, pp. 10\,283--10\,294, 2020.

\bibitem{Hind-2016}
H.~Albasry and Q.~Z. Ahmed, ``Network-assisted d2d discovery method by using
  efficient power control strategy,'' in \emph{2016 IEEE 83rd Vehicular
  Technology Conference (VTC Spring)}, 2016, pp. 1--5.

\bibitem{Fuhu-2023}
F.~Che, Q.~Z. Ahmed, F.~A. Khan, and F.~A. Khan, ``Novel fine-tuned attribute
  weighted naïve bayes {NLoS} classifier for {UWB} positioning,'' \emph{IEEE
  Communications Letters}, vol.~27, no.~4, pp. 1130--1134, 2023.

\bibitem{Ahmed-2008}
Q.~Z. Ahmed, W.~Liu, and L.-L. Yang, ``Least mean square aided adaptive
  detection in hybrid direct-sequence time-hopping ultrawide bandwidth
  systems,'' in \emph{VTC Spring 2008 - IEEE Vehicular Technology Conference},
  2008, pp. 1062--1066.

\bibitem{Ahmed-2014}
Q.~Z. Ahmed, K.-H. Park, M.-S. Alouini, and S.~Aïssa, ``Compression and
  combining based on channel shortening and reduced-rank techniques for
  cooperative wireless sensor networks,'' \emph{IEEE Transactions on Vehicular
  Technology}, vol.~63, no.~1, pp. 72--81, 2014.

\bibitem{Wang2020_CSI}
X.~Wang, X.~Wang, and S.~Mao, ``Deep convolutional neural networks for indoor
  localization with {CSI} images,'' \emph{{IEEE} Transactions on Network
  Science and Engineering}, vol.~7, pp. 316--327, 1 2020.

\bibitem{Ryan-2017}
R.~Husbands, Q.~Ahmed, and J.~Wang, ``Transmit antenna selection for massive
  mimo: A knapsack problem formulation,'' in \emph{2017 IEEE International
  Conference on Communications (ICC)}, 2017, pp. 1--6.

\bibitem{Amjad-2023}
B.~Amjad, Q.~Z. Ahmed, P.~I. Lazaridis, F.~A. Khan, M.~Hafeez, and Z.~D.
  Zaharis, ``Deep learning approach for optimal localization using an mm-wave
  sensor,'' \emph{IEEE Transactions on Instrumentation and Measurement},
  vol.~72, pp. 1--15, 2023.

\bibitem{Osama-2017}
O.~Alluhaibi, Q.~Z. Ahmed, C.~Pan, and H.~Zhu, ``Hybrid digital-to-analog
  beamforming approaches to maximise the capacity of mm-wave systems,'' in
  \emph{2017 IEEE 85th Vehicular Technology Conference (VTC Spring)}, 2017, pp.
  1--5.

\bibitem{Nair-2016}
M.~Nair, Q.~Z. Ahmed, and H.~Zhu, ``Hybrid digital-to-analog beamforming for
  millimeter-wave systems with high user density,'' in \emph{2016 IEEE Global
  Communications Conference (GLOBECOM)}, 2016, pp. 1--6.

\bibitem{iGentUWBAoA}
M.~Naseri, A.~Shahid, G.-J. Gordebeke, S.~Lemey, M.~Boes, S.~Van De~Velde, and
  E.~De~Poorter, ``Machine learning-based angle of arrival estimation for
  ultra-wide band radios,'' \emph{IEEE Communications Letters}, vol.~26, no.~6,
  pp. 1273--1277, 2022.

\bibitem{Chen2020_IMU}
C.~Chen, P.~Zhao, C.~X. Lu, W.~Wang, A.~Markham, and N.~Trigoni,
  ``Deep-learning-based pedestrian inertial navigation: Methods, data set, and
  on-device inference,'' \emph{{IEEE} Internet of Things Journal}, vol.~7, pp.
  4431--4441, 5 2020.

\bibitem{Schwarzbach2022}
P.~Schwarzbach, R.~Weber, and O.~Michler, ``Statistical evaluation and
  synthetic generation of ultra-wideband distance measurements for indoor
  positioning systems,'' \emph{IEEE Sensors Journal}, vol.~22, no.~6, pp.
  4836--4843, 2022.

\bibitem{DS_Girolami}
M.~Girolami, F.~Furfari, P.~Barsocchi, and F.~Mavilia, ``A bluetooth 5.1
  dataset based on angle of arrival and {RSS} for indoor localization,''
  \emph{IEEE Access}, vol.~11, pp. 81\,763--81\,776, 2023.

\bibitem{DS_Maus}
\BIBentryALTinterwordspacing
G.~Maus, H.~Pörner, R.~Schlenke, R.~Ahrens, S.~Janicke, P.~Bolz, and
  E.~Dürholt, ``Bluetooth 5.1 angle of arrival based indoor localization,''
  2021. [Online]. Available: \url{https://dx.doi.org/10.21227/2j4h-3w77}
\BIBentrySTDinterwordspacing

\bibitem{Maus_respiration}
G.~Maus and D.~Brückmann, ``Joint angle and respiration estimation for passive
  and device-free respiration monitoring,'' in \emph{ICASSP 2023 - 2023 IEEE
  International Conference on Acoustics, Speech and Signal Processing
  (ICASSP)}, 2023, pp. 1--5.

\bibitem{Maus_localization}
G.~Maus, H.~Pörner, R.~Ahrens, and D.~Brückmann, ``A phase normalization
  scheme for angle of arrival based bluetooth indoor localization,'' in
  \emph{2022 IEEE 65th International Midwest Symposium on Circuits and Systems
  (MWSCAS)}, 2022, pp. 1--5.

\bibitem{DS_Piazzese}
\BIBentryALTinterwordspacing
N.~I. Piazzese, M.~Perrone, and D.~P. Pau, ``Dataset for bluetooth 5.1
  direction of arrival with non uniform rectangular arrays,'' \emph{Data in
  Brief}, vol.~39, p. 107576, 2021. [Online]. Available:
  \url{https://www.sciencedirect.com/science/article/pii/S2352340921008519}
\BIBentrySTDinterwordspacing

\bibitem{DS_Paulino}
\BIBentryALTinterwordspacing
N.~Paulino, ``A dataset of phase samples using an 8-element uniform circular
  antenna array and a bluetooth low energy 5.1 nordic {nRF52811} based
  receiver,'' 2022. [Online]. Available:
  \url{https://dx.doi.org/10.21227/92ba-e365}
\BIBentrySTDinterwordspacing

\bibitem{paper_Paulino}
N.~Paulino and L.~M. Pessoa, ``Self-localization via circular bluetooth 5.1
  antenna array receiver,'' \emph{IEEE Access}, vol.~11, pp. 365--395, 2023.

\bibitem{Ahmed-2020}
Q.~Z. Ahmed, M.~Hafeez, F.~A. Khan, and P.~Lazaridis, ``Towards beyond 5g
  future wireless networks with focus towards indoor localization,'' in
  \emph{2020 IEEE Eighth International Conference on Communications and
  Networking (ComNet)}, 2020, pp. 1--5.

\bibitem{Leitch-2023}
\BIBentryALTinterwordspacing
S.~G. Leitch, Q.~Z. Ahmed, W.~B. Abbas, M.~Hafeez, P.~I. Laziridis,
  P.~Sureephong, and T.~Alade, ``On indoor localization using {WiFi}, {BLE},
  {UWB}, and {IMU} technologies,'' \emph{Sensors}, vol.~23, no.~20, 2023.
  [Online]. Available: \url{https://www.mdpi.com/1424-8220/23/20/8598}
\BIBentrySTDinterwordspacing

\bibitem{ble5.3_spec}
\BIBentryALTinterwordspacing
(1999) Bluetooth core specification bluetooth. {Vol}. 6, Part B pp. 2726-2727.
  (accessed March 28, 2023). [Online]. Available:
  \url{https://www.bluetooth.com/specifications/specs/core-specification-5-3/}
\BIBentrySTDinterwordspacing

\bibitem{Mohaghegh2021}
P.~Mohaghegh, A.~Boegli, and Y.~Perriard, ``Bluetooth low energy direction
  finding principle,'' in \emph{ICEMS 2021 - 2021 24th International Conference
  on Electrical Machines and Systems}.\hskip 1em plus 0.5em minus 0.4em\relax
  Institute of Electrical and Electronics Engineers Inc., 2021, pp. 830--834.

\bibitem{rtls_toolbox}
\BIBentryALTinterwordspacing
T.~I. Incorporated. (2019) {RTLS Toolbox}. (accessed December 18, 2023).
  [Online]. Available:
  \url{https://software-dl.ti.com/simplelink/esd/simplelink_cc13x2_26x2_sdk/4.10.00.78/exports/docs/ble5stack/ble_user_guide/html/ble-stack-5.x-guide/localization-index-cc13x2_26x2.html#}
\BIBentrySTDinterwordspacing

\bibitem{IIoTLab}
\BIBentryALTinterwordspacing
U.~of~Ghent. (2023) Industrial {IoT} lab. (accessed December 12, 2023).
  [Online]. Available:
  \url{https://www.ugent.be/ea/idlab/en/research/research-infrastructure/industrial-iot-lab.htm}
\BIBentrySTDinterwordspacing

\bibitem{diff_geom}
\BIBentryALTinterwordspacing
A.~Manikas, \emph{Differential Geometry in Array Processing}.\hskip 1em plus
  0.5em minus 0.4em\relax Imperial College Press, 2004. [Online]. Available:
  \url{https://skynet.ee.ic.ac.uk/ambook/2004_Diff_Geometry_Manikas_1860944221.pdf}
\BIBentrySTDinterwordspacing

\bibitem{qualisys_rotations}
\BIBentryALTinterwordspacing
E.~Schoonderwaldt and D.~Thompson. (2016, 5) Learn about 6{DOF}. (accessed
  December 13, 2023). [Online]. Available:
  \url{https://www.qualisys.com/webinars/learn-about-6dof/}
\BIBentrySTDinterwordspacing

\bibitem{aoa_orientation}
F.~Mavilia, P.~Barsocchi, F.~Furfari, D.~La~Rosa, and M.~Girolami, ``On the
  analysis of body orientation for indoor positioning with {BLE} 5.1 direction
  finding,'' in \emph{ICC 2023 - IEEE International Conference on
  Communications}, 2023, pp. 204--209.

\bibitem{crlb_ble}
W.~Shi, B.~Huang, and K.~Sun, ``A {CRLB} analysis of {AoA} estimation using
  bluetooth 5,'' in \emph{ICASSP 2022 - 2022 IEEE International Conference on
  Acoustics, Speech and Signal Processing (ICASSP)}, 2022, pp. 5158--5162.

\bibitem{Zafari2019}
F.~Zafari, A.~Gkelias, and K.~K. Leung, ``A survey of indoor localization
  systems and technologies,'' \emph{IEEE Communications Surveys and Tutorials},
  vol.~21, pp. 2568--2599, 2019.

\bibitem{Leitch_nb}
S.~Leitch, Q.~Z. Ahmed, P.~Lazaridis, and M.~Hafeez, ``Exploiting channel
  diversity to improve {BLE} range-finding accuracy,'' in \emph{Proceedings of
  the UNIfied Conference of DAMAS, IncoME and TEPEN Conferences (UNIfied
  2023)}, A.~D. Ball, H.~Ouyang, J.~K. Sinha, and Z.~Wang, Eds.\hskip 1em plus
  0.5em minus 0.4em\relax Cham: Springer Nature Switzerland, 2024, pp.
  591--599.

\bibitem{Nikodem2021}
M.~Nikodem and P.~Szelinski, ``Channel diversity for indoor localization using
  bluetooth low energy and extended advertisements,'' \emph{IEEE Access},
  vol.~9, pp. 169\,261--169\,269, 2021.

\bibitem{Ahmed-2013}
Q.~Z. Ahmed, M.-S. Alouini, and S.~Aissa, ``Bit error-rate minimizing detector
  for amplify-and-forward relaying systems using generalized gaussian kernel,''
  \emph{IEEE Signal Processing Letters}, vol.~20, no.~1, pp. 55--58, 2013.

\bibitem{gpr_jadaliha}
M.~Jadaliha, Y.~Xu, J.~Choi, N.~S. Johnson, and W.~Li, ``Gaussian process
  regression for sensor networks under localization uncertainty,'' \emph{IEEE
  Transactions on Signal Processing}, vol.~61, no.~2, pp. 223--237, 2013.

\bibitem{gpr_esteve}
E.~Aparicio-Esteve, W.~Raes, N.~Stevens, J.~Ure\~{n}a, and A.~Hern\'{a}ndez,
  ``Experimental evaluation of a machine learning-based rss localization method
  using gaussian processes and a quadrant photodiode,'' \emph{Journal of
  Lightwave Technology}, vol.~40, no.~19, pp. 6388--6396, 2022.

\bibitem{gpr_book}
\BIBentryALTinterwordspacing
C.~E. Rasmussen and C.~K.~I. Williams, \emph{{Gaussian Processes for Machine
  Learning}}.\hskip 1em plus 0.5em minus 0.4em\relax The MIT Press, 11 2005.
  [Online]. Available: \url{https://doi.org/10.7551/mitpress/3206.001.0001}
\BIBentrySTDinterwordspacing

\bibitem{Hensen-2013}
J.~Hensman, N.~Fusi, and N.~D. Lawrence, ``Gaussian processes for big data,''
  in \emph{Proceedings of the Twenty-Ninth Conference on Uncertainty in
  Artificial Intelligence}, ser. UAI'13.\hskip 1em plus 0.5em minus 0.4em\relax
  Arlington, Virginia, USA: AUAI Press, 2013, p. 282–290.

\end{thebibliography}

\end{document}